\documentclass[conference]{IEEEtran}
\usepackage{xcolor}
\makeatletter
\def\footnoterule{\relax%
  \kern-5pt
  \hbox to \columnwidth{\hfill\vrule width 0.8\columnwidth height 0.4pt\hfill}
  \kern4.6pt}
\makeatother
\usepackage{textcomp}

\usepackage{cite}

\ifCLASSINFOpdf
   \usepackage[pdftex]{graphicx}
   \graphicspath{{../pdf/}{../jpeg/}}
   \DeclareGraphicsExtensions{.pdf,.jpeg,.png}
\else
   \usepackage[dvips]{graphicx}
   \graphicspath{{../eps/}}
   \DeclareGraphicsExtensions{.eps}
\fi

\usepackage{balance}

\usepackage{siunitx} 

\usepackage{multirow}

\usepackage{xcolor}
\definecolor{redcolor}{rgb}{1, 0, 0}

\usepackage{subfiles}
\usepackage{makecell}

\IEEEoverridecommandlockouts
\begin{document}
\setlength{\textfloatsep}{5pt}
%

\title{Reservoir Computing with Planar\\ Nanomagnet Arrays\\
}


\author{
    \IEEEauthorblockN{Peng Zhou\IEEEauthorrefmark{1}, Nathan R. McDonald\IEEEauthorrefmark{2}, Alexander J. Edwards\IEEEauthorrefmark{1}, Lisa Loomis\IEEEauthorrefmark{2}, Clare D. Thiem\IEEEauthorrefmark{2}, Joseph S. Friedman\IEEEauthorrefmark{1}}
    \IEEEauthorblockA{\IEEEauthorrefmark{1}Department of Electrical and Computer Engineering, The University of Texas at Dallas, Richardson, TX
    \\\{peng.zhou, alexander.edwards, joseph.friedman\}@utdallas.edu}
    \IEEEauthorblockA{\IEEEauthorrefmark{2}Air Force Research Laboratory - Information Directorate, Rome, NY
    \\\{nathan.mcdonald.5, lisa.loomis.3, clare.thiem\}@us.af.mil}
}

%
%
%
%

%


\maketitle


\begin{abstract}

Reservoir computing is an emerging methodology for neuromorphic computing that is especially well-suited for hardware implementations in size, weight, and power (SWaP) constrained environments.  This work proposes a novel hardware implementation of a reservoir computer using a planar nanomagnet array.  A small nanomagnet reservoir is demonstrated via micromagnetic simulations to be able to identify simple waveforms with 100\% accuracy.  Planar nanomagnet reservoirs are a promising new solution to the growing need for dedicated neuromorphic hardware.
\\
\end{abstract}
\renewcommand\IEEEkeywordsname{Keywords}

%

\section{Introduction}
\begin{figure}[t]
    \centering
    \includegraphics[width=250pt]{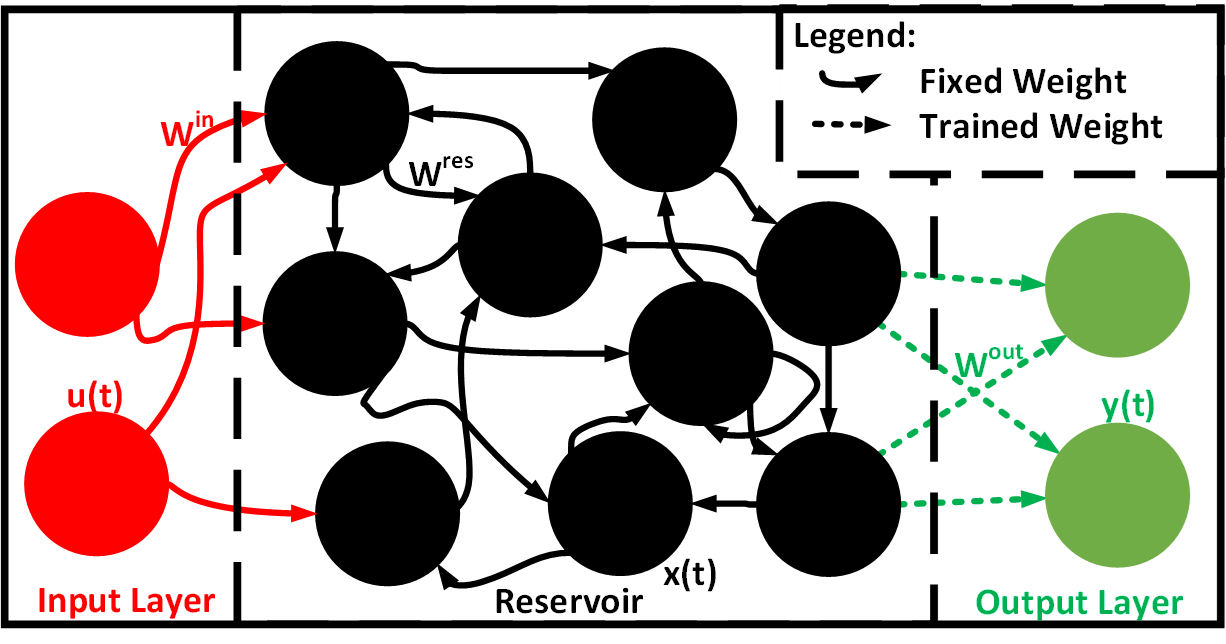}
    \caption{A generic reservoir computer.
    (adapted from \cite{Ganguly17})
    }
    \label{fig:RCDiag}
\end{figure}

Recurrent neural networks (RNN) are neural networks (NN) that allow for feedback loops within the NN. Unlike conventional feedforward NNs that use only the current inputs, RNNs use current inputs in conjunction with the history of previous states to determine the network output. Reservoir computing (RC) \cite{Jaeger01, Maass02, Jaeger04, Tanaka19} is a subset type of RNN, where, after weight initialization, only the weights of the output layer are altered during training (Fig. \ref{fig:RCDiag}). This alternate training scheme does not require multi-layer backpropagation and gradient descent, making RC simpler to train. Various software and mathematical models of RCs have demonstrated that this simplified network approach has not diminished its computing capability, competing with conventional deep neural networks in spatio-temporal data analysis \cite{Tanaka19, Ganguly17}. Further still, direct hardware implementations promise to be significantly more SWaP-efficient, since expensive circuitry for updating internal reservoir weights is unnecessary. This technique is therefore well suited for resource constrained hardware environments.

Recently, spintronic devices exhibiting nanoscale magnetic phenomena have been developed that provide opportunities for applications in computing, especially in RC. Magnetic memory devices using magnetic tunnel junctions (MTJs) \cite{Kawahara12} are already produced in industry, and there is now much interest in using arrays of nanomagnets to compute and propagate binary information through nanomagnetic logic (NML) \cite{Imre06, Niemier12, Alawein19, Turvani17, Jensen18}. Memory and logic solutions with spintronics require these complex physical devices to operate within a digital paradigm, drastically constraining their operation. Without these constraints, however, some spintronic devices can exhibit highly nonlinear behavior that is extremely well-suited for implementing RC in hardware.

Various hardware RC implementations have been proposed \cite{Jensen18, Loomis18, Furuta18, Nomura19, Hassan17, Du17}, but they all require external stimuli to propagate information within the reservoir or maintain the reservoir state. The solution presented here, using non-volatile planar nanomagnet arrays, propagates information passively, allowing surrounding circuitry to be extremely simple and power-efficient.

\begin{figure}[t]
    \centering
    \includegraphics[width=205pt]{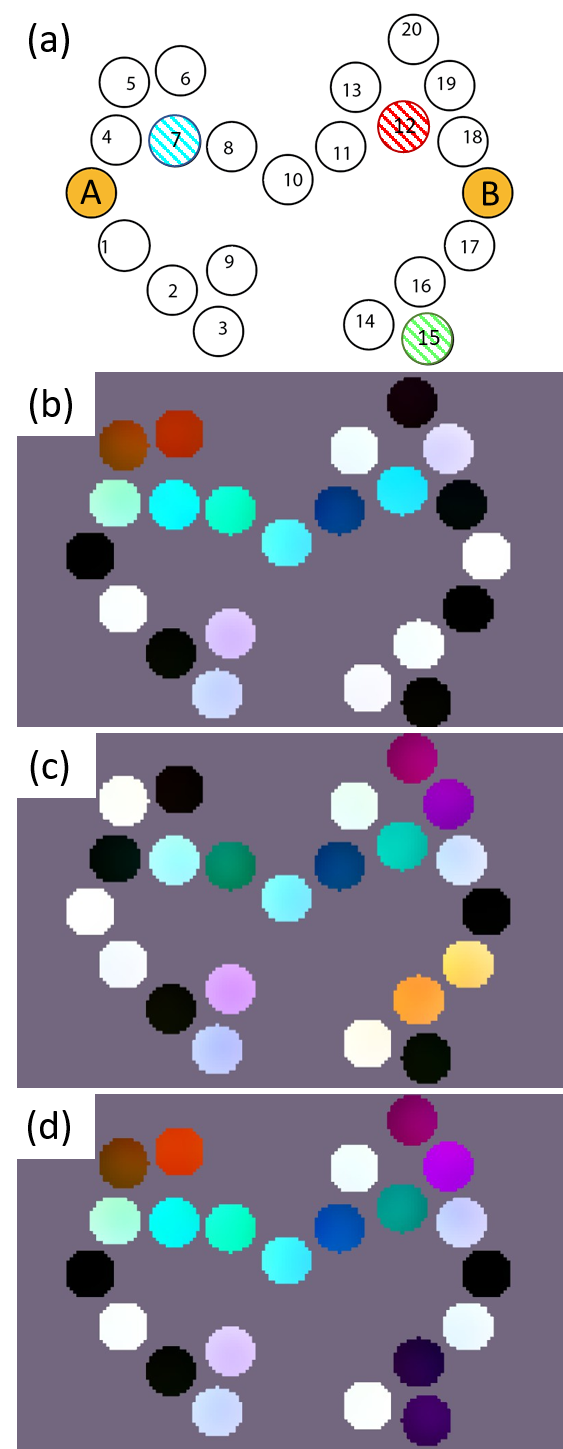}
    \caption{\textbf{(a)} Reservoir layout: Nanomagnets A and B are the inputs to the reservoir, while the highlighted nanomagnets are those indicated in Fig \ref{fig:wavesTotal}(b).  \textbf{(b-d)} Micromagnetic simulation snapshots of the network at $t = 63$, $66$, and $69$ ns.  Lighter (darker) colors of nanomagnets indicate magnetization in the positive (negative) $z$ direction.  The magnetization of each nanomagnet varies over time, indicating information is propagating among the nanomagnets.}
    \label{fig:networkDiag}
\end{figure}

\begin{figure*}[t]
    \centering
    \includegraphics[width=530pt]{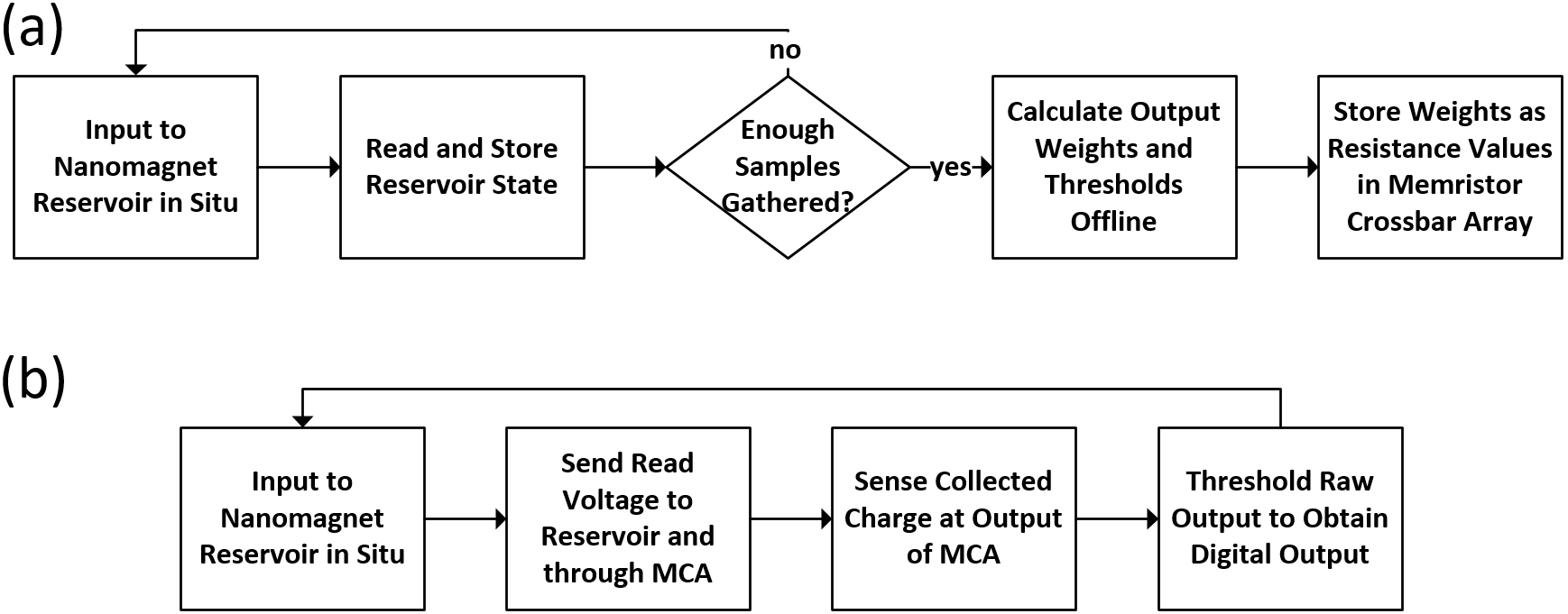}
    \caption{\textbf{(a)} Training flow: Information is presented to the reservoir by forcing the magnetizations of specific nanomagnets.  The other nanomagnets react to decrease the total energy of the system.  The states of these nanomagnets are recorded after every input.  After enough samples are gathered, they are processed along with the expected output.  The output weights are determined via ridge regression and recorded in the resistance states of the memristors.  \textbf{(b)}  Inference flow: Inputs are presented to the reservoir as in the training flow.  The sampled reservoir states are fed directly into the MCA which performs vector-matrix multiplication to obtain an output vector of currents.  These currents are sensed to obtain the final digital output vector.}
    \label{fig:flowchart}
\end{figure*}

\begin{figure}[t]
    \centering
    \includegraphics[width=250pt]{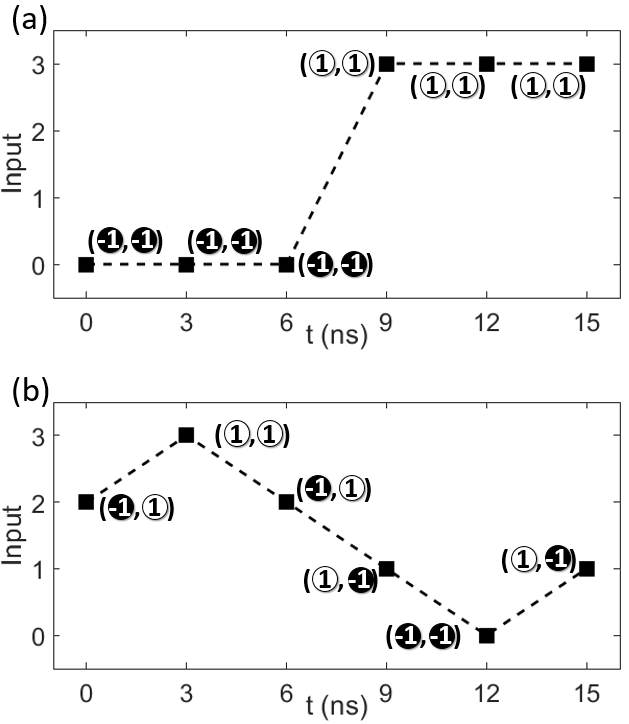}
    \caption{ Input encoding scheme for the \textbf{(a)} square-wave and \textbf{(b)} triangle-wave.  White (black) represents forced magnetization in the positive (negative) $z$-direction.  Sets of six inputs of either a square or triangle wave were randomly concatenated together to form the input string.}
    \label{fig:encoding}
\end{figure}

\begin{figure*}[t]
    \centering
    \includegraphics[width=530pt]{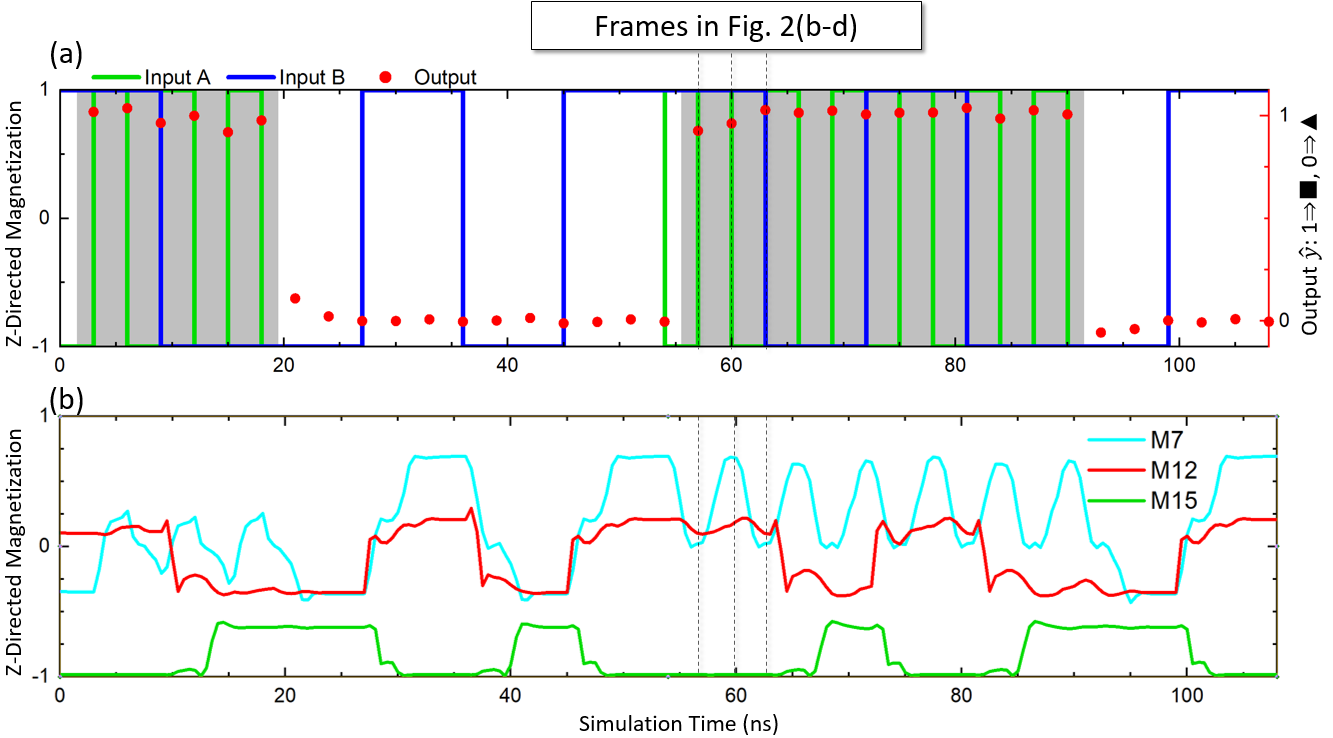}
    \caption{\textbf{(a)} Representative inputs to the reservoir:  Gray (white) regions of the plot indicate a triangle (square) wave.  The calculated outputs $\mathbf{\hat{y}}(t)$ are shown as red dots, where a linear classifier (threshold 0.5) is the final step.  The samples highlighted by the vertical black lines are the frames depicted in Fig. \ref{fig:networkDiag}(b-d). \textbf{(b)} Plot of magnetization over time for three nanomagnets: Nanomagnet 7 tracks input A, nanomagnet 12 tracks input B, and nanomagnet 15 is affected by both inputs through nonlinear reservoir node interactions.}
    \label{fig:wavesTotal}
\end{figure*}

\section{Reservoir Computing}

\subsection{Theory}

The internal dynamics of the reservoir layer state are described by:

\begin{equation}
    \mathbf{x}(t) = f(\mathbf{W^{in}u}(t), \mathbf{W^{res}x}(t-\tau)),
    \label{eq:x}
\end{equation}

\noindent where $\mathbf{u}(t)$ are the inputs, $\mathbf{W^{in}}$ are the input weights, $\mathbf{x}(t)$ are the reservoir states, $\mathbf{W^{res}}$ are the reservoir weights, and $\tau$ is the sample period  \cite{Jaeger01, Maass02}. Given the desired output state of the reservoir $\mathbf{y}(t)$, the necessary output weights $\mathbf{W^{out}}$ can be calculated via ridge regression:

\begin{equation}
    \mathbf{W^{out}} = \mathbf{yx}^\top (\mathbf{xx}^\top+\lambda \mathbf{I})^{-1},
    \label{eq:Wout}
\end{equation}

\noindent where $\top$ denotes the transpose, $\lambda$ is the regularization factor, and $\mathbf{I}$ is the identity matrix. During testing, the actual reservoir output $\mathbf{\hat{y}}(t)$ is calculated as 

\begin{equation}
    \mathbf{\hat{y}}(t) = \mathbf{W^{out}x}(t).
    \label{eq:yhat}
\end{equation}

\subsection{Nanomagnet Reservoir}

This work proposes a novel reservoir composed of a planar arrangement of nanomagnets each having perpendicular magnetic anisotropy (PMA). Fig. \ref{fig:networkDiag}(a) shows an example layout of nanomagnets. The magnetization of each nanomagnet produces a magnetic field that induces a change in the magnetizations of the other nanomagnets. The effect of these magnetic fields upon nanomagnet magnetization is sigmoidal in shape due to the anisotropy \cite{Cha17}, and a nanomagnet\textquotesingle s ability to influence another varies with the distance between the two. These two features, non-linear interaction and variable interaction strength, make the proposed implementation well-suited for neuromorphic computing.
The distances between nanomagnets cannot be changed after fabrication (i.e. the network weights cannot be updated), so the network cannot be trained conventionally. However, if these nodes and weights are considered the reservoir layer, then these weights do not need to be changed at all, making planar nanomagnet arrays a good match for RC.

Information is input by stimulating individual nanomagnets or portions of the nanomagnet reservoir with external magnetic fields. The magnetodynamics within the reservoir propagate the input information non-linearly throughout the reservoir. The $z$ components of the magnetizations of various nanomagnets are read electrically via MTJs. A complementary single layer circuit is used to perform vector-matrix multiplication on the magnetization values and output weights to obtain the output vector. Only this final layer needs to be adjusted to train the neural network, which may be performed with a memristor crossbar array (MCA) with current sense amplifiers at the output \cite{Alibart13}. Once the output weights are trained, the reservoir can tested against the given task. Fig. \ref{fig:flowchart} depicts a flowchart describing the training and inference methodologies. Because of the compact design, passive information propagation within the reservoir, and non-volatile nature of the reservoir, the proposed nanomagnet implementation is ideal for SWaP-constrained environments.

\section{Waveform Identification}

\subsection{Micromagnetic Simulation Methodology}

A reservoir layer based on planar nanomagnet arrays is simulated and trained with a simple waveform identification task. At each time step, the RC is tasked to determine whether a pair of inputs are from a square or triangle wave. These waves are quantized to two bits \cite{Loomis18}, and there is a 3 ns delay between each input $\mathbf{u}(t)$.  This is sufficient time for the nanomagnets to relax towards stable states.  Each bit is fed into the reservoir simultaneously by forcing the magnetization in the + (1) or - (0) $z$-direction (Fig. \ref{fig:encoding}).

Immediately before every input, the $z$ component of the magnetization of each nanomagnet is recorded as the reservoir state, $\mathbf{x}(t)$. At the end of the simulation, the output weights, $\mathbf{W^{out}}$, are calculated in a single calculation step, Eq. (\ref{eq:Wout}). The output weights are then used to generate the output, $\mathbf{\hat{y}}(t)$, of the network at every time step in accordance with Eq. (\ref{eq:yhat}).

Micromagnetic simulations are performed with MuMax3 \cite{Vansteenkiste2014} for the nanomagnet network pictured in Fig. \ref{fig:networkDiag}(a) using Table \ref{tab:params} parameters. The nanomagnet dynamics are simulated with a time step on the order of femtoseconds to obtain physically accurate results. This network has two input nodes and 20 reservoir nodes, all of which exhibit PMA. All of the reservoir nodes were treated as outputs.

\newcommand{\minus}{\scalebox{0.75}[1.0]{$-$}}
\renewcommand\theadalign{bc}
\renewcommand\theadfont{\bfseries}
\renewcommand\theadgape{\Gape[4pt]}
\renewcommand\cellgape{\Gape[4pt]}

\begin{table}
    \centering
    \caption{\label{tab:table-name}  Simulation Parameters}
    \begin{tabular}{|c|c|c|c|}
    \hline
        \thead{Parameter} & \thead{Description} & \thead{Value} & \thead{Unit}\\
        \hline
        $Msat$ & Saturation Magnetization & $7.23e5$ & $A/m$\\
        \hline
        $Aex$ & Exchange Stiffness & $1.3e\minus11$ & $J/m$\\
        \hline
        $\alpha$ & Gilbert Damping Factor & $0.5$ & \\
        \hline
        $Ku_i$ & Input Anisotropy & $3.62e5$ & $J/m^3$\\
        \hline
        $Ku_r$ & Reservoir Anisotropy & $1.05e5$ & $J/m^3$\\
        \hline
        $D$ & Nanomagnet Diameter & $30e\minus9$ & $m$\\
        \hline
        $th$ & Nanomagnet Thickness & $12e\minus9$ & $m$\\
        \hline
        $T$ & Input Period & $3e\minus9$ & $s$\\
        \hline
    \end{tabular}
    \label{tab:params}
\end{table}

\subsection{Waveform Identification Results}

The nanomagnet reservoir was simulated with an input stream comprised of 25 triangle or square wave periods, with 6 input pairs per wave, for a total of 150 input pairs.  120 of these were used to train the output weights and the remaining 30 were used to test the RC. The reservoir identified the waveforms with 100\% accuracy for both the training and testing data. Fig. \ref{fig:networkDiag}(b-d) shows various snapshots of the network during the simulation. Fig. \ref{fig:wavesTotal} shows a segment of the inputs and outputs to and from the network, along with waveforms of the magnetizations of various nanomagnets. The planar nanomagnet array thus successfully performed a simple RC task.

\section{Conclusion}

RC is a methodology for neuromorphic computing well-suited for dedicated hardware implementations in SWaP-constrained environments. This work presents a hardware implementation of an RC using a planar nanomagnet array that is demonstrated to identify simple waveforms with 100\% accuracy. Further work will be done to prove that these reservoirs will be successful on more complex tasks with many inputs. Passive planar nanomagnet arrays are therefore a promising solution for dedicated neuromorphic hardware.

\section*{Acknowledgement}

Effort sponsored by the Air Force under contract number FA8750-15-3-6000. The U.S. Government is authorized to reproduce and distribute copies for Governmental purposes notwithstanding any copyright or other restrictive legends. The views and conclusions contained herein are those of the authors and should not be interpreted as necessarily representing the official policies or endorsements, either expressed or implied, of the Air Force or the U.S. Government. The authors thank E. Laws, J. McConnell, N. Nazir, L. Philoon, and C. Simmons for technical support, and the Texas Advanced Computing Center at The University of Texas at Austin for providing computational resources.  Distribution A.  Approved for public release; distribution is unlimited. 88ABW-2020-0939



\addtolength{\textheight}{-12cm} 

\bibliographystyle{IEEEtran}
\balance
\bibliography{IEEEabrv,main}

\begin{thebibliography}{10}
\providecommand{\url}[1]{#1}
\csname url@samestyle\endcsname
\providecommand{\newblock}{\relax}
\providecommand{\bibinfo}[2]{#2}
\providecommand{\BIBentrySTDinterwordspacing}{\spaceskip=0pt\relax}
\providecommand{\BIBentryALTinterwordstretchfactor}{4}
\providecommand{\BIBentryALTinterwordspacing}{\spaceskip=\fontdimen2\font plus
\BIBentryALTinterwordstretchfactor\fontdimen3\font minus
  \fontdimen4\font\relax}
\providecommand{\BIBforeignlanguage}[2]{{%
\expandafter\ifx\csname l@#1\endcsname\relax
\typeout{** WARNING: IEEEtran.bst: No hyphenation pattern has been}%
\typeout{** loaded for the language `#1'. Using the pattern for}%
\typeout{** the default language instead.}%
\else
\language=\csname l@#1\endcsname
\fi
#2}}
\providecommand{\BIBdecl}{\relax}
\BIBdecl

\bibitem{Jaeger01}
H.~Jaeger, ``The" echo state" approach to analysing and training recurrent
  neural networks-with an erratum note','' \emph{Bonn, Germany: German National
  Research Center for Information Technology GMD Technical Report}, vol. 148,
  01 2001.

\bibitem{Maass02}
W.~Maass \emph{et~al.}, ``Real-time computing without stable states: a new
  framework for neural computation based on perturbations.'' \emph{Neural
  Comput.}, vol.~14, no.~11, 2002.

\bibitem{Jaeger04}
H.~Jaeger \emph{et~al.}, ``Harnessing nonlinearity: Predicting chaotic systems
  and saving energy in wireless communication,'' \emph{Science}, vol. 304,
  no.~78, 2004.

\bibitem{Tanaka19}
G.~Tanaka \emph{et~al.}, ``Recent advances in physical reservoir computing: A
  review,'' \emph{Neural Networks}, vol. 115, 2019.

\bibitem{Ganguly17}
S.~Ganguly \emph{et~al.}, ``Reservoir computing using stochastic p-bits,''
  \emph{arXiv: 1709.10211}, 2017.

\bibitem{Kawahara12}
T.~Kawahara \emph{et~al.}, ``{Spin-transfer torque RAM technology: Review and
  prospect},'' pp. 613--627, apr 2012.

\bibitem{Imre06}
A.~Imre \emph{et~al.}, ``Majority logic gate for magnetic quantum-dot cellular
  automata,'' \emph{Science}, vol. 311, no. 5758, 2006.

\bibitem{Niemier12}
M.~T. Niemier \emph{et~al.}, ``Shape engineering for controlled switching with
  nanomagnet logic,'' \emph{IEEE Trans. Nanotechnol.}, vol.~11, no.~2, 2012.

\bibitem{Alawein19}
M.~Alawein \emph{et~al.}, ``Multistate nanomagnet logic using equilateral
  permalloy triangles,'' \emph{IEEE Magn. Lett.}, vol.~10, 2019.

\bibitem{Turvani17}
G.~Turvani \emph{et~al.}, ``{A pNML Compact Model Enabling the Exploration of
  Three-Dimensional Architectures},'' \emph{IEEE Transactions on
  Nanotechnology}, vol.~16, no.~3, pp. 431--438, may 2017.

\bibitem{Jensen18}
J.~H. Jensen \emph{et~al.}, ``Computation in artificial spin ice,'' \emph{The
  2019 Conference on Artificial Life}, no.~30, 2018.

\bibitem{Loomis18}
L.~Loomis \emph{et~al.}, ``An {FPGA} implementation of a time delay reservoir
  using stochastic logic,'' \emph{J. Emer. Tech. Comp.}, vol.~14, no.~4, 2018.

\bibitem{Furuta18}
T.~Furuta \emph{et~al.}, ``Macromagnetic simulation for reservoir computing
  utilizing spin dynamics in magnetic tunnel junctions,'' \emph{Phys. Rev.
  App.}, vol.~10, 2018.

\bibitem{Nomura19}
H.~Nomura \emph{et~al.}, ``Reservoir computing with dipole-coupled
  nanomagnets,'' \emph{Jap. J. App. Phys.}, vol.~58, no.~7, 2019.

\bibitem{Hassan17}
A.~M. Hassan \emph{et~al.}, ``Hardware implementation of echo state networks
  using memristor double crossbar arrays,'' in \emph{IJCNN}, 2017, pp.
  2171--2177.

\bibitem{Du17}
C.~Du \emph{et~al.}, ``Reservoir computing using dynamic memristors for
  temporal information processing,'' \emph{Nat Commun.}, vol.~8, no.~1, 2017.

\bibitem{Cha17}
I.~H. Cha \emph{et~al.}, ``Perpendicular magnetic anisotropy and interfacial
  dzyaloshinskii-moriya interaction in pt/cofesib structures,'' \emph{IEEE
  Magn. Lett.}, vol.~8, 2017.

\bibitem{Alibart13}
F.~Alibart \emph{et~al.}, ``{Pattern classification by memristive crossbar
  circuits using ex situ and in situ training},'' \emph{Nat Commun.}, vol.~4,
  no.~1, 2013.

\bibitem{Vansteenkiste2014}
A.~Vansteenkiste \emph{et~al.}, ``{The design and verification of Mumax3},''
  \emph{AIP Adv.}, vol.~4, no.~10, p. 107133, 2014.

\end{thebibliography}

\end{document}